\documentclass[ twoside,openright,titlepage,numbers=noenddot,
                headinclude,footinclude,cleardoublepage=empty,abstract=on,
                BCOR=5mm,paper=a4,fontsize=11pt
                ]{scrartcl}
\usepackage{graphicx}
\usepackage{microtype}

\begin{document}

\title{Bridging the Gap Between Explainable AI and Uncertainty Quantification to Enhance Trustability
}
\subtitle{A Call to Action\\}


\author{Dominik Seuss
}




\begin{center}

    \begingroup
        \large \textls[160]{BRIDGING THE GAP BETWEEN EXPLAINABLE AI AND UNCERTAINTY QUANTIFICATION TO ENHANCE TRUSTABILITY}\\
        \vspace{10pt}
        A Call to Action
    \endgroup
\end{center}
\vspace{20pt}
Dominik Seuss \\
\textit{Fraunhofer Institute for Integrated Circuits IIS, Erlangen, Germany}

\section*{Abstract}
After the tremendous advances of deep learning and other AI methods, more attention is flowing into other properties of modern approaches, such as interpretability, fairness, etc. combined in frameworks like Responsible AI. Two research directions, namely Explainable AI and Uncertainty Quantification are becoming more and more important, but have been so far never combined and jointly explored. In this paper, I show how both research areas provide potential for combination, why more research should be done in this direction and how this would lead to an increase in trustability in AI systems.
\paragraph{Keywords:} Responsible AI, Trust, Explainable AI, Uncertainty Quantification

\section{Introduction}
\label{intro}

Deep Learning methods have achieved tremendous success in almost all disciplines in the field of Artificial Intelligence and Machine Learning. Now, there are signs of a shift away from the goal of improving the recognition performance of the models alone. Frameworks like ``Responsible AI'' are now coming to the fore, postulating properties like fairness, interpretability, privacy and security of AI applications.
In this paper, I will focus on the notion of interpretability in particular, expand interpretability to include ``Trust'' and show why more research should go into combining the methods of Explainable AI and Uncertainty Quantification.
To this end, I will take up these two separate research branches, point out their methods, and show their relatedness in contributing towards trustability \-- the quality or state of being trustable.
There are already research fields such as ``Planning under Uncertainty'' and ``Decision making under Uncertainty'' which combine known methods with uncertainty but these either assume that uncertainties already exist or refer to methods for quantifying uncertainty \cite{blythe1999overview,dimitrakakis2018decision} without having properties such as trustability towards humans as a goal.

No one would question that explainability of approaches is conducive to building trust. But uncertainties and awareness of them also have an impact on people. A number of behavioral and electrophysiological studies indicate ``[...] a strong relationship between uncertainty and a key component of cognitive control \-- outcome monitoring. In particular, it appears that highly uncertain environments tend to increase the recruitment of monitoring processes'' \cite{mushtaq2011uncertainty}. So the more uncertain a situation is for us, the more we try to monitor the outcome. Applied to AI systems, this would mean a stronger need for explanation and predictability of the outcome. Both properties that can be achieved through a combination of Explainable AI and Uncertainty Quantification methods.
As the coordinator for ``Explainable Learning'', one of the main research focus areas at the ADA Lovelace Center for Analytics, Data and Applications\footnote{https://www.scs.fraunhofer.de/en/focus-projects/ada-center.html}, I bring those researchers working on explainability of AI approaches together with those working on Uncertainty Quantification.

In the next section, both branches of research with their definitions and methods will be introduced. In the third section, possible combinations of the two research directions and the importance of their combination will be discussed. The outlook will be a suggestion on how to handle trust in the future.

\section{Overview of Explainable AI and Uncertainty Quantification}
\label{sec:2}

\subsection{Explainable AI}
\label{subsec:2.1}
\begin{figure*}
\center
  \includegraphics[scale=0.2]{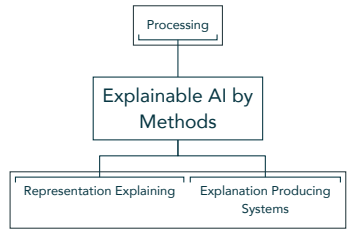}
\caption{Overview of Explainable AI by Methods. All three categories (Processing, Representation Explaining, and Explanation Producing Systems) can be further divided into post-explainability (upper box) and explainable modelling (lower box).}
\label{fig:XAI_Overview}       
\end{figure*}

There is no clear definition of Explainable AI. The Defense Advanced Research Projects Agency (DARPA, \cite{darpa}) formulates the goals as ``produce more explainable models, while maintaining a high level of learning performance (prediction accuracy); and enable human users to understand, appropriately, trust, and effectively manage the emerging generation of artificially intelligent partners''. Another definition comes from the organisers of xML Challenge, FICO \cite{fico}, who see XAI as ``an innovation towards opening up the black-box of ML'' and as ``a challenge to create models and techniques that are both accurate and provide good trustworthy explanation that will satisfy customers' needs''. Both definitions already contain the term ``trust'' without going further into it.
An overview of the individual methods without concrete approaches can be seen in Figure \ref{fig:XAI_Overview} (for an in-depth survey please refer to \cite{adadi2018peeking}). 

One way to group Explainable AI methods is to divide them into three categories according to their chronological order of use: ``Pre-modelling explainability'', ``Explainable Modelling'' and ``Post-modelling Explainability'' \cite{towards}. We will leave out the ``Pre-modelling explainability'' category for this paper, as this deals with understanding the data itself before actually training a model.

\begin{itemize}
	\item Explainable modelling\\
For explainable modelling, a component for comprehensibility is already implemented in the model during training. Two types of approaches can be distinguished: ``Representation Explaining'' and ``Explanation Producing Systems'' (see lower block in Figure \ref{fig:XAI_Overview}). An example for Representation Explaining are Concept Activation Vectors \cite{kim2018interpretability}, which can be used to show for example how sensitive a prediction of ``zebra'' is to the presence of stripes. Approaches like Beta VAE \cite{Higgins2017betaVAELB}, from the field of Explanation Producing Systems, try to introduce interpretable latent representations, which means that the latent representations contain human-interpretable information already.
	\item Post-Explainability\\
In Post-Explainability, after the decision of the model, the processing is examined up to the decision (see upper block in Figure \ref{fig:XAI_Overview}). The different approaches use different types of and representations for explanations. An example for a graphical representation is ``Layerwise Relevance Propagation'' \cite{lrp} and is often used with images and shows the user which pixels of the input made what contribution to the decision. An example for a different type of representation is another approach named ANN-DT \cite{809084}. It extracts decision trees from neural networks to make the underlying decision processes more comprehensible for humans.
\end{itemize}

\subsection{Uncertainty Quantification}
\label{subsec:2.2}
One comprehensive description \cite{defUQ} of Uncertainty Quantification (UQ) from the field of applied mathematics is: 
``UQ involves the quantitative characterisation and management of uncertainty in a broad range of applications. It employs both computational models and observational data, together with theoretical analysis. UQ encompasses many different tasks, including uncertainty propagation, sensitivity analysis, statistical inference and model calibration, decision making under uncertainty, experimental design, and model validation. UQ thus draws upon many foundational ideas and techniques in applied mathematics and statistics (e.g., approximation theory, error estimation, stochastic modelling, and Monte Carlo methods) but focuses these techniques on complex models, for instance, of physical or socio-technical systems that are primarily accessible through computational simulation. [...]''

\begin{figure*}
\center
  \includegraphics[scale=0.2]{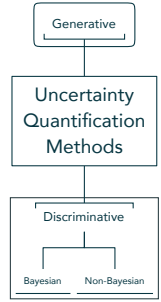}
\caption{Overview of Uncertainty Quantification Methods in Machine Learning}
\label{fig:UQ_Overview}       
\end{figure*}

This definition also holds for the field of machine learning, since we do almost always deal with complex models. There are two main sources of uncertainty, aleatoric uncertainty and epistemic uncertainty \cite{hullermeier2020aleatoric}. The former is caused by noise in data or labels, and the latter is caused by data sparsity or out of data distribution.
The scope of this paper is not to give a comprehensive overview of uncertainty quantification methods, but to provide a categorisation of selected approaches (see Figure \ref{fig:UQ_Overview}) to show their combinability with Explainable AI methods. For a more extensive treatise on uncertainty quantification methods, see \cite{abdar2020review}. 
In AI research, Uncertainty Quantification can be used to pursue two goals. First, to calibrate neural networks so that the output confidence reflects the empirical accuracy, and second, for out-of-distribution detection, i.e., to detect whether a new classification example really corresponds to the distribution with which the network was trained.

Uncertainty quantification methods can be divided into two types, discriminative methods and generative methods. Generative methods are used, among other things, to perform out-of-distribution detection \cite{grathwohl2020classifier}, or to reconstruct data, such as image reconstruction from noisy and incomplete images \cite{bohm2019uncertainty}.
Discriminative methods - the focus of this work - are used to perform e.g. classification, i.e. to assign an input to one or more or no output classes. Here, one can distinguish between model-agnostic approaches, Bayesian and Non-Bayesian methods.

\begin{itemize}
	\item Model-agnostic methods\\
This includes approaches that are independent of the method used. They generate e.g. through data augmentation, better calibrated DNNs (Mixup \cite{thulasidasan2019mixup}) or alleviate the over-confidence of the model (CutMix \cite{yun2019cutmix}). 
	\item Bayesian methods\\
They use Bayesian Inference, in which Bayes' theorem is used to update the probability for a hypothesis as more evidence or information becomes available. Two well-known approaches of this type are Bayes by Backprop \cite{pmlr-v37-blundell15} and MC Dropout \cite{pmlr-v48-gal16}.
	\item Non-bayesian methods\\
These include approaches that are not based on Bayesian inference. Reasons for this are, e.g., that Bayesian approaches require significant modifications to the training procedure and are computationally expensive compared to non-Bayesian neural networks. One well-known approach is to estimate uncertainty using deep ensembles \cite{lakshminarayanan2017simple}. In this context, the quantification of Uncertainty comes from sampling from several models, where each model is trained separately on the same data or subsets. The Consensus of the models gives confidence in overall model ensemble.
\end{itemize} 

\section{Discussion}
In this section, I will address the question of why more research should be done in the direction of joint approaches for Explainable AI and Uncertainty Quantification and will also discuss the desired properties of such an approach. Then two short exemplary possible combinations of methods from Explainable AI and Quantification of Uncertainties will be shown. Finally, I will discuss the concept of trust and highlight why both research areas can jointly contribute to improving trustability.
\subsection{Combination of Explainable AI and Uncertainty Quantification}
\label{sec:3.1}
When one speaks of Interpretability, one automatically thinks of Explainable AI methods. The research field of Uncertainty Quantification is almost not present here. In my opinion, however, current research needs to be more concerned with merging the two research directions, as in a sense they show two sides of the same coin. While the methods of Explainable AI try to show the way to the decision, the methods of Quantification of Uncertainty try to give a realistic evaluation regarding the reliability of the decision. Thus, through their respective statements, ``I know why I made a decision and can show it to you'' and ``I know what I don't know or how certain I am about my decision'', both research directions contribute to paint a more complete picture and thus should end up in their combination being more than the sum of their parts.

Sections~\ref{subsec:2.1} and \ref{subsec:2.2} have shown different methods that differ in their complexity and in their integrability, but can form a basis for combinations.
On a meta-level, the result of such a combination should be a model that combines both Explainable AI and Uncertainty Quantification, so that one gets back not only an explanation but also an uncertainty in the decision including an indication of where the uncertainty comes from. This must be done in a way that is appropriate for humans:
The nature of the explanations must be both understandable to humans and delivered in an appropriate manner.
A few possibilities for this are:
\begin{itemize}
	\item Concrete explanations: The user is directly provided with the explanation in the form of decision $\rightarrow$ certainty $\rightarrow$ explanation. An example of this in an emotion classification task would be the feedback ``The person is classified as sad to 30\% due to a covering of the mouth.''
	\item Counterfactual Explanations: The user is shown under which conditions the system would have decided differently and with what degree of certainty. Applied to the example just presented, the feedback would be ``If the person's mouth was not covered, I would classify the person as 80\% happy''. User-studies \cite{psych1,dodge2019explaining} have shown, for example, that people prefer counterfactual explanations over case-based reasoning, i.e., solving new problems based on the solutions of similar past problems. \newline
\end{itemize}

The type of representation also has to be understandable to humans. It is very input-specific and must be appropriate and expectable for the user depending on the problem. Some forms of representation that could be considered are:
\begin{itemize}
	\item Textual\\
The system provides an explanation in the form of a text in which the uncertainty of the decision is also verbalised. For example, a combination of approaches for generating interpretable latent representations, presented in section~\ref{subsec:2.1} and ``Bayes by Backprop'', presented in section~\ref{subsec:2.2}, would be conceivable. 
	\item Visual\\
The system provides an explanation in the form of a visualisation with an appropriate representation of uncertainties. One possibility here would be, for example, to link the method LRP, presented in section~\ref{subsec:2.1}, with an elliptic coding of uncertainties obtained by Monte Carlo dropout procedure, presented in section~\ref{subsec:2.2}.
\end{itemize}

\subsection{The Role of Trust}
\label{sec:3.2}
Early in the introduction, I talked about trust and its psychological role and necessity to humans. In other disciplines, especially in human-machine interaction, trust has played a major role from the very beginning. Parts of these definitions from other research branches can also be transferred to the requirements of Responsible AI.

In robotics, Lewis et al. \cite{lewis2018role} distinguish two properties of systems: System predictability and system intelligibility and transparency. The former is important because ``[...] knowing that the automation may fail reduces the uncertainty and consequent risk associated with use of the automation. In other words, predictability may be as (or more) important as reliability.'' This in combination with ``Systems that can explain their reasoning will be more likely to be trusted, since they would be more easily understood by their users [...]'' describe in my opinion exactly this combination of Explainable AI (``Systems that can explain their reasoning [...]'') and Uncertainty Quantification (``[...] predictability may be as (or more) important as reliability''). 

A system that can tell the user that it is uncertain in combination with a reasoning for it corresponds to my propositions from section~\ref{sec:3.1} and should have as a goal an enhancement of trustability in AI systems.

\section{Outlook}
It is remarkable that both definitions of Explainable AI in section~\ref{subsec:2.1} address trust, even if it only seems more like trust is a pleasant side effect of explainability. At this point, I wonder if this has led to less attention being paid to linking it to quantification of uncertainty, on the one hand, and whether this classification is appropriate, on the other hand. It would also be conceivable to rename the ``Interpretability'' pillar in the Responsible AI framework to ``Trust'' and to downgrade Explainable AI to a method for increasing the trustability of an AI system. Even if the actual idea behind the point ``Interpretability'' is perhaps to check whether AI systems function as intended, this goal can also be subordinated to the term ``Trust'' or trust-building measure. Perhaps it will also come about in the near future that one will focus even more on the human component and human acceptance in this case and carry out such a renaming.

\subsubsection*{Acknowledgements}
The author wants to thank Teena Hassan for the helpful discussions.

%
%

\bibliographystyle{spmpsci}      
\bibliography{bib.bib}   

%
%

\end{document}